\documentclass[12pt,a4paper]{article}
\usepackage[utf8]{inputenc}
\usepackage[T1]{fontenc}
\usepackage{amsmath}
\usepackage{amsfonts}
\usepackage{fancyhdr}
\usepackage{amssymb}
\usepackage{color}
\definecolor{Prune}{RGB}{99,0,60}
\usepackage{mdframed}
\usepackage{multirow}
\usepackage{multicol}
\usepackage{scrextend}
\usepackage{tikz}
\usepackage{graphicx}
\usepackage[absolute]{textpos} 
\usepackage{colortbl}
\usepackage{array}
\usepackage{geometry}
\usepackage{hyperref}
\usepackage{algorithm}
\usepackage{algpseudocode}
\usepackage{makecell}

\title{Memory Networks: Towards Fully Biologically Plausible Learning}
\author{
    Jacobo Ruiz$^{1,2}$\footnote{Technical report submitted in partial fulfillment of the requirements for a Master’s degree. Work conducted while attached to A*STAR.}, Manas Gupta$^{1}$ \\
    $^1$ Institute for Infocomm Research (I\textsuperscript{2}R), A*STAR, Singapore \\
    $^2$ Paris-Saclay University, France
}
\date{}

\begin{document}

\maketitle

\begin{abstract}
The field of artificial intelligence faces significant challenges in achieving both biological plausibility and computational efficiency, particularly in visual learning tasks. Current artificial neural networks, such as convolutional neural networks, rely on techniques like backpropagation and weight sharing, which do not align with the brain’s natural information processing methods. To address these issues, we propose the Memory Network, a model inspired by biological principles that avoids backpropagation and convolutions, and operates in a single pass. This approach enables rapid and efficient learning, mimicking the brain’s ability to adapt quickly with minimal exposure to data. Our experiments demonstrate that the Memory Network achieves efficient and biologically plausible learning, showing strong performance on simpler datasets like MNIST. However, further refinement is needed for the model to handle more complex datasets such as CIFAR-10, highlighting the need to develop new algorithms and techniques that closely align with biological processes while maintaining computational efficiency.
\end{abstract}

\section{Introduction}

The pursuit of algorithmic efficiency in artificial intelligence is increasingly drawing inspiration from the brain's remarkable capabilities. The human brain, with its approximately 20 watts of power consumption \cite{attwell2001energy}, exemplifies an unparalleled level of efficiency in managing and processing information, especially when compared to the staggering energy demands of modern AI models. For instance, training a large-scale model like GPT-3 \cite{Brown2020} is estimated to require 1,287 MWh of electricity \cite{AIIndex2023}—a quantity of energy that the human brain would consume over approximately 7,346 years. This stark contrast highlights a critical need to explore and approximate the efficiency of the brain in our computational systems, particularly in areas like visual learning where the parallels between biological and artificial systems are most evident.\vspace{0.2cm}

Despite their inspirations from the brain, current deep learning models, such as convolutional neural networks (CNNs) \cite{LeCun1998} and other deep networks, exhibit key differences from their biological counterparts. These models are typically trained using backpropagation—a method that, while effective, lacks biological plausibility due to its reliance on global error signals and non-local computations. Biological neural networks, by contrast, learn through more localized and distributed processes which do not require global coordination or precise error feedback mechanisms.

The significance of bridging this gap between biological and artificial systems is underscored by the potential gains in computational efficiency. The brain's ability to process complex sensory information and adapt through plasticity mechanisms, without relying on backpropagation, suggests that biologically inspired approaches could yield more efficient and scalable AI models \cite{lake2017building}. This is particularly relevant in visual learning tasks, where the hierarchical organization \cite{hubel1962receptive} and local connectivity of neurons in the visual cortex offer a compelling blueprint for developing more energy-efficient and adaptive algorithms.\vspace{0.2cm}

Biological plausibility in neural network design entails several criteria: local computation, where neurons update based on nearby signals rather than global information \cite{Douglas1991}; no backpropagation of error signals, avoiding the need for non-local error propagation \cite{Crick1989}; 
 unsupervised or semi-supervised learning \cite{Olshausen1996}, which mirrors the brain's ability to learn from unstructured data without explicit labels; and rapid learning on a physiologically limited setup.\vspace{0.2cm}

Some approaches have been developed to make modern models more biologically plausible by addressing specific aspects of this challenge. For example, alternatives to the backpropagation algorithm \cite{Rumelhart1986}, such as Hebbian learning, have been inspired by the neurobiology of neuronal function \cite{Hebb1949}. However, these models still only partially adhere to biological principles. For instance they rely on CNNs which use convolutions to mimic the brain’s local receptive fields and hierarchical feature learning, but these convolutions rely on non-biological practices like weight sharing. \vspace{0.2cm}

To address these limitations, we propose the Memory Network, an approach designed to more closely adhere to the principles of biological plausibility.
During training, the model encodes data representations by updating each neuron based on the similarity between a new input and the average representations of previous inputs with the same label. This single-pass learning approach allows the model to quickly learn and store patterns, mimicking the brain’s ability to adapt rapidly with minimal exposure. During prediction, the test input is compared to these learned representations, and the neuron with the closest match determines the output.\\
The Memory Network operates without backpropagation, relying instead on local plasticity mechanisms for learning, which aligns more closely with biological processes observed in the brain. By avoiding convolutions and traditional feature extraction methods, the model enhances its biological plausibility. It supports semi-supervised learning by depending only partially on labels, reflecting the brain’s unsupervised learning capabilities. Moreover, the model’s architecture is simpler than traditional neural networks, allowing it to train significantly faster without the need for extensive preprocessing techniques. This efficiency is largely due to the fact that training is completed in a single epoch, with each data point processed only once, setting it apart from other biologically plausible methods that require multiple passes through the data or complex preprocessing steps.\vspace{0.2cm}

In summary, the Memory Network model represents an initiative to explore how to make AI systems more biologically plausible by adhering to principles based on biology. This approach paves the way for further research in developing AI models that operate more like the brain, promoting advancements in energy-efficient and biologically inspired computing.

\section{Related Work}

\subsection{Criteria for Biological Plausibility}

Biological plausibility in neural network design involves several key principles that ensure these models more closely mimic the brain’s natural processes. One important criterion is local computation, where neurons or units update their states and adjust synaptic weights based only on information available in their immediate vicinity. This mirrors the way biological neurons operate, with synaptic changes driven by local signals like neurotransmitter release rather than signals from the entire network \cite{Douglas1991}.

Another crucial aspect is avoiding the propagation of global error signals across the entire network. Backpropagation, a foundational algorithm in deep learning \cite{Rumelhart1986}, is widely regarded as not biologically plausible due to several key discrepancies between its operation in artificial neural networks (ANNs) and how learning occurs in biological neural networks. In biological systems, learning is governed by local mechanisms, where neurons adjust their synaptic strengths based on local activity and signals, without the need to propagate error gradients backward through the network. In contrast, backpropagation requires non-local information, such as error signals that traverse multiple layers, and assumes precise symmetry of weights between neurons during the forward and backward passes. These requirements are not found in biological neural networks, where synaptic changes are driven by local processes and lack such symmetric and globally coordinated weight updates \cite{Crick1989}.

Furthermore, biological learning often happens without explicit supervision, relying on unsupervised or semi-supervised learning principles to interpret unstructured data. This approach mirrors the brain’s remarkable ability to adapt and learn from its environment without requiring explicit labels or instructions \cite{Olshausen1996}.

In addition to these principles, single epoch-based learning is an important feature, reflecting the brain’s capacity for rapid learning from limited exposure \cite{Standing1973}.  The brain often learns and adapts in real-time, making swift adjustments to new information without the need for repeated trials. This fast-paced learning and real-time processing and inference are essential for survival and effective functioning in dynamic environments.

\subsection{Previous Work on Biologically Plausible Algorithms}

A variety of studies have sought to develop learning algorithms that are more aligned with neurobiological principles, aiming to create models that better mimic the mechanisms observed in the brain. One prominent approach is Hebbian learning, based on the principle that formulated by Donald O. Hebb as "neurons that fire together wire together" \cite{Hebb1949}. This principle has been effectively implemented in models like those by Miconi et al., where a Hebbian network was used before a Convolutional Neural Network, achieving around 60\% accuracy on the CIFAR-10 dataset \cite{miconi2021hebbianlearninggradientshebbian}\cite{Krizhevsky2009}. Beyond Hebbian learning, other biologically inspired algorithms, such as Feedback Alignment, Direct Feedback Alignment, Difference Target Propagation, and Predictive Coding, have been explored as potential biologically plausible alternatives to backpropagation \cite{Lillicrap2016}\cite{Nokland2016}\cite{Lee2015}\cite{Rao1999}. Hebbian learning has been shown to outperform backpropagation in accuracy in some cases, and most biologically inspired methods converge faster than backpropagation \cite{gupta2023bioinspiredlearningbetterbackprop}.\vspace{0.2cm}

However, many of these approaches, despite their grounding in neurobiology, often incorporate elements like convolutions or rely on backpropagation to achieve high performance \cite{Moraitis_2022} \cite{lagani2022fasthebbscalinghebbiantraining} . For example, in "Hebbian Learning Meets Deep Convolutional Neural Networks," Amato et al. used a Hebbian-based method as a feature extractor in conjunction with a Convolutional Neural Network, reaching approximately 80\% accuracy on CIFAR-10 \cite{inbook}. \vspace{0.2cm}

Another promising approach in biologically plausible computation is Spiking Neural Networks (SNNs). SNNs emulate the natural information processing of the brain by using discrete spikes, or action potentials, for neuron communication, in contrast to the continuous activations used in traditional artificial neural networks. These networks incorporate time-dependent processes and more closely mimic the dynamics of biological neurons, offering a more realistic model for neural computation \cite{MAASS19971659}. However, achieving high performance (98\%-99\% accuracy on MNIST) with SNNs often requires adaptations to backpropagation-like algorithms \cite{hunsberger2015training} \cite{rueckauer2017conversion}.\vspace{0.2cm}

\begin{figure}[h]
    \centering
    \includegraphics[width=\textwidth]{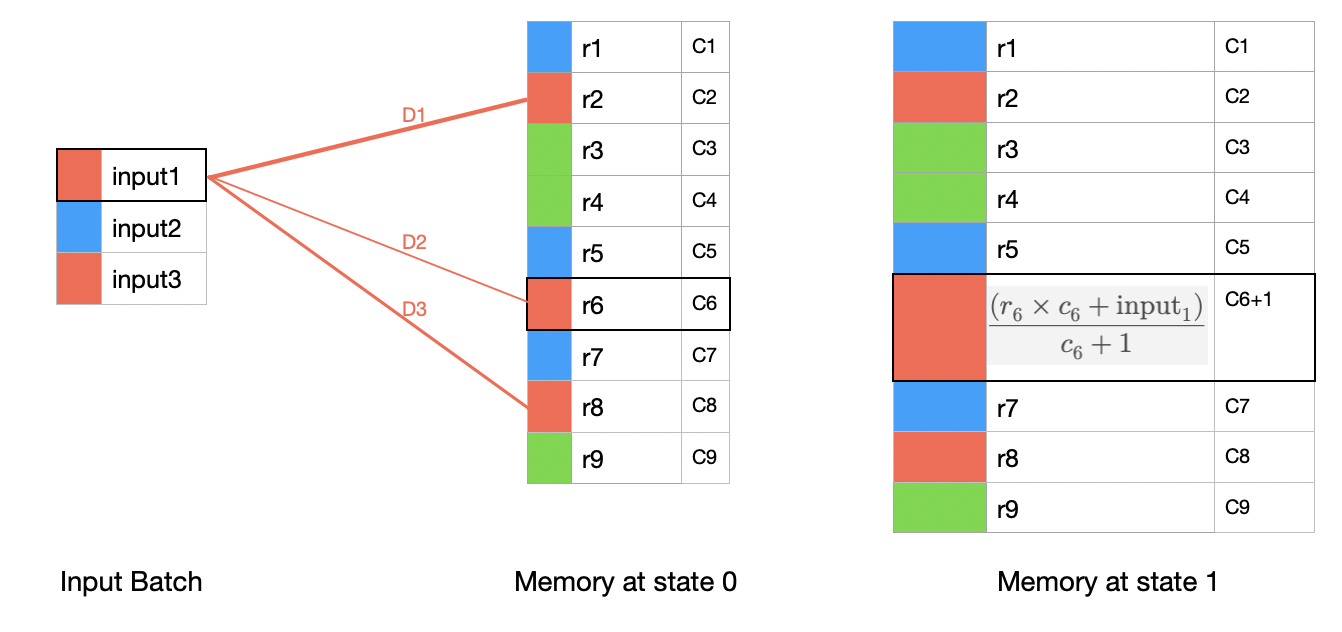}
    \caption{\textbf{Update Mechanism with Full Memory:} In this simplified example, the closest neuron to input1 is determined based on the computed distances (D1, D2, D3) to neurons with the same color label at memory state 0. The smallest distance, D2, indicates that input1 is closest to the neuron in row 6, leading to the latent representation r6 being updated with input1 and counter c6, as shown in memory state 1. This update process is conducted in parallel for the entire batch.}
    \label{fig:memupdate2}
\end{figure}

\subsection{Similar algorithms to Memory Networks}
The proposed algorithm, called the Memory Network, diverges from the aforementioned methods as it does not use backpropagation or convolutions. Unlike other biologically inspired models that serve primarily as feature extractors for subsequent processing stages, our Memory Network is designed as a self-contained, biologically plausible solution for visual classification tasks. It operates by rapidly processing data in a single pass, more closely mirroring how learning and memory function in biological neural systems.\vspace{0.2cm}

In addition to deep learning methods, alternative algorithms that do not rely on backpropagation or convolutions can complete training in a single epoch. Algorithms like K-means clustering \cite{Lloyd1982} and K-Nearest Neighbors (KNN) \cite{Cover1967} have been explored for tasks such as image classification on datasets like CIFAR-10. However, only KNN will be considered here, as K-means performs only slightly better than random chance on CIFAR-10-like classification tasks \cite{pmlr-v15-coates11a}.\vspace{0.2cm}

A key difference between our Memory Network and KNN lies in memory usage. KNN requires storing all training points in memory, while our Memory Network undergoes a training phase that allows for storing a more compact set of representations. Algorithmically, our model has a training phase where our model learns the training set representations unlike KNN which does not perform training set processing.\vspace{0.2cm}

While k-means and KNN can converge quickly and work well under limited computational resources, their accuracy is generally lower than that of deep learning models, especially without feature extraction or dimensionality reduction. For example, KNN with Principal Component Analysis (PCA) \cite{pearson1901pca} achieves an accuracy of about 41.78\% on CIFAR-10, which is comparable to our highest accuracy \cite{abouelnaga2016cifar10knnbasedensembleclassifiers}.\vspace{0.2cm}

However, techniques like PCA, often used with KNN to reduce data dimensionality, are not biologically plausible and can be computationally expensive, especially with relatively large datasets like CIFAR-10. The brain relies on non-linear, local, and real-time processing mechanisms that are continuously adaptive, whereas PCA is a linear, global, batch-oriented algorithm that lacks these characteristics \cite{Olshausen1996} \cite{pmlr-v15-coates11a}.\vspace{0.2cm}

This approach does not necessarily offer a more time-efficient solution and relies on methods that do not align with biological plausibility principles. Our Memory Network, by contrast, avoids such preprocessing steps and provides a more biologically aligned approach that performs similarly.

\section{Methodology}
\subsection{Biological Motivation and Design Choices}

The design of the algorithm and its components were inspired by biological neural processes:

\begin{itemize}
    \item \textbf{Single Epoch Based Learning}: The model is trained in a single pass through the data, reducing the need for repeated exposure to the same data, emulating how the brain quickly adapts and learns from new experiences with minimal repetition.

    \item \textbf{Dynamic Adaptation and Learning}: The use of counters and latent representations allows neurons to dynamically adjust based on the frequency of input exposure, much like how biological neurons modify their synaptic strengths according to activity levels (Hebbian learning). 

    \item \textbf{Simple Network Structure}: The algorithm employs a multi-neuron structure with no layers, which is reminiscent of the initial layers of neural processing in the brain that handle basic feature extraction.

    \item \textbf{Fast Processing and Inference}: The network is designed to train and infer faster than deep neural networks, aligning with the brain’s capability for rapid learning and decision-making.
\end{itemize}

\subsection{Algorithm Overview}

The algorithm introduced in this methodology is inspired by biological neural processes and designed to efficiently learn and adapt in both supervised and unsupervised settings. It utilizes a memory-based approach, where each neuron stores its own label, latent representation, and a counter tracking its exposure to inputs. The training process involves dynamically assigning inputs to neurons, either by initializing unassigned neurons or by selecting the closest matching neuron based on a computed distance. This enables the model to incrementally learn and adapt its internal representations in response to incoming data.
\\

\textbf{Pseudocode Representation}:  
The pseudocode provided represents a simplified sequential version of the actual implementation, focusing on the core logic of the training and prediction processes.
 
\begin{algorithm}
\caption{Simplified Memory Network Algorithm}
\begin{algorithmic}[1]
\State // Training function:
\State \textbf{Input:} $\text{max\_neurons}$, \text{training\_batch}
\State \text{// Initialize empty memory}
\State Initialization:
\State $N \gets  \text{max\_neurons} \text{ empty neurons}$
\State $\text{neuron\_counter} \gets 0$
\For{each $(x, y)$ in \text{training\_batch}}
    \If{ $N$ is full}
        \State // Get the neurons with the same label as the input
        \State $\text{neurons\_with\_same\_label} \gets \{n \in N \mid n.\text{label} = y\}$
        \State 
        \State // Compute the distances between the input and each selected neuron
        \State $\text{distances} \gets \text{DistanceFunction}(\text{neurons\_with\_same\_label}, x)$
        \State 
        \State // Find the minimum distance and the closest neuron
        \State $\text{min\_distance}, \text{closest\_neuron} \gets \min(\text{distances})$
        \State
        \State // Memory update: Assign input to the closest neuron
        \State $N[\text{closest\_neuron}].\text{update\_representation}(x)$
        \State $N[\text{closest\_neuron}].\text{update\_counter}(+1)$

    \Else
        \State // Memory update: Assign input and label to an empty neuron
        \State // Get the next available empty neuron
        \State $\text{new\_neuron} \gets N[\text{neuron\_counter}]$
        \State $N[\text{new\_neuron}].\text{update\_representation}(x)$
        \State $N[\text{new\_neuron}].\text{update\_label}(y)$
        \State $N[\text{new\_neuron}].\text{update\_counter}(+1)$
        \State $\text{neuron\_counter} \gets \text{neuron\_counter} + 1$
    \EndIf
\EndFor
\State // Prediction function:
\State \textbf{Input:} $\text{max\_neurons}$, \text{test\_batch}
\State \text{// Initialize empty memory}
\State Initialization:
\State $N \gets  \text{max\_neurons} \text{ neurons with learned representations}$
\State $\text{predictions} \gets \text{empty list}$
\For{each $x$ in \text{test\_batch}}
    \State // Compute the distances between the input and all neurons
    \State $\text{distances} \gets \text{DistanceFunction}(N, x)$
    \State 
    \State // Find the minimum distance and the closest neuron
    \State $\text{min\_distance}, \text{closest\_neuron} \gets \min(\text{distances})$
    \State
    \State // Get the label of the closest neuron and add it to predictions
    \State $\text{predicted\_label} \gets N[\text{closest\_neuron}].\text{label}$
    \State $\text{predictions.append(predicted\_label)}$
\EndFor
\State \Return $\text{predictions}$

\end{algorithmic}
\end{algorithm}

\subsection{Central Memory Structure: Neuron Tensor}

A key component of the algorithm is the central memory structure, referred to as the \textit{Neuron Tensor}. This tensor is designed to store essential information about each neuron in the model. Each row in the tensor corresponds to a neuron and contains three pieces of information:
\begin{itemize}
    \item \textbf{Label}: Indicates the class that the neuron represents.
    \item \textbf{Latent Representation}: A learned vector representation that captures the neuron’s characteristics or "memory" of the inputs assigned to it. Specifically it is a tensor of the same shape as the input. 
    \item \textbf{Counter}: Tracks the number of inputs assigned to each neuron, enabling dynamic adjustment of neuron properties based on exposure frequency.
\end{itemize}

This design choice is inspired by biological neurons, which adapt based on the frequency and type of inputs they receive. In our model, the counter serves a similar purpose by tracking the adaptation process and influencing the neuron's response to new inputs.\\

All the neurons are initialized to an "unintialized" or "empty" state, meaning the latent space is a tensor of zeros, the counter is initialized to 0 and the label is set to -1.

\subsection{Processing of a training batch and neuron assignment}

In this subsection, we describe how a training batch is processed by assigning each input to a corresponding neuron.

Two techniques are described, one using thresholds and one not using thresholds.

\subsubsection{Basic technique}
This process occurs in two scenarios: (1) when neurons are still uninitialized, and (2) when all neurons are initialized (non-zero tensors, counters > 0, labels not -1).

In the first scenario, with uninitialized neurons, each input is assigned directly to an uninitialized neuron without distance computation. Once all neurons are initialized, any remaining inputs follow the procedure for a fully initialized memory.

In the second scenario, when neurons are initialized (Figure \ref{fig:memupdate2}), inputs are assigned based on label and distance. Inputs are matched with neurons of the same label, selecting the closest neuron for memory updates.

Initially, this process relies on input labels with uninitialized neurons. Once fully initialized, the method can switch to unsupervised training by assigning inputs to the closest neuron, ignoring labels.

\subsubsection{Technique using a Threshold}
We developed a thresholding technique to improve the representations learnt by the neurons. This technique has two scenarios: one for when the memory is not full and another for when it is full. In the second scenario, the method behaves like the basic approach. In the first scenario, there are some key differences.\vspace{0.2cm}

In scenario 1, when a new input arrives, we calculate the distances between the input and the existing neurons of the same label, as we do when the memory is full. However, if the minimum distance exceeds a certain threshold, the input is assigned to an empty neuron. If the distance is below this threshold, the input is assigned to the closest neuron. Once the memory becomes full, the basic technique is applied.\vspace{0.2cm}

The threshold value is based on the full centroids per class, computed from the entire training set. This approach means that these threshold values are determined prior to training.

The centroid for each class \(k\) is calculated as follows:

\begin{equation}
\mathbf{c}_k = \frac{1}{N_k} \sum_{i=1}^{N_k} \mathbf{x}_i
\end{equation}

where \(\mathbf{c}_k\) is the centroid of class \(k\), \(N_k\) is the number of samples in class \(k\), and \(\mathbf{x}_i\) represents each sample of class \(k\) in the training set.

This centroid is used to compute the average distance between samples of class \(k\) and the centroid itself. The idea is to estimate the expected average distance between the class representation and its samples. During training, if a distance greatly exceeds this average, the input is considered a new representation and assigned to a new neuron. This helps improve the representations learnt, by categorizing representations very different from those seen previously in training, as new neurons. \vspace{0.2cm}

To fine-tune the parameters, a coefficient is added to this centroid value. Thus, there is one centroid per class and one coefficient per image channel.
 
\subsection{Memory Update Step}

This section explains how a neuron's latent representation is updated when an input is assigned to it:

\[
l_{n+1} = \frac{l_n \cdot n + \text{input}}{n + 1}
\]

where \(l_n\) is the current average representation, \(n\) is the count of inputs so far, and \(\text{input}\) is the new value. This unsupervised update (as no label is used or error is computed) incorporates the new input into the average. After updating, \(n\) is incremented to \(n + 1\).
\subsection{Distance Function Formulation}

To determine the similarity between input images and the neurons, the algorithm employs a distance function. Two variations of this function were explored:

\begin{itemize}
    \item \textbf{Absolute Value of Errors (Manhattan Distance)}:
    \[
    D_{\text{abs}}(x, y) = \sum_{i} |x_i - y_i|
    \]
    This measure computes the absolute differences between corresponding elements of the input vector \(x\) and the neuron representation \(y\). It reflects the total variation between the two, similar to how the brain might process differences in sensory input in a straightforward, linear manner.

    \item \textbf{Euclidean Distance}:
    \[
    D_{\text{euclid}}(x, y) = \sqrt{\sum_{i} (x_i - y_i)^2}
    \]
    The Euclidean distance considers the geometric distance in a multi-dimensional space, providing a more nuanced measure that takes into account both the magnitude and direction of the differences. This is akin to the brain's ability to compute more complex patterns, not just linear differences.
\end{itemize}

For images with \(c\) channels (e.g., RGB with 3 channels), the distance between an input and a neuron is computed by first calculating the distance for each channel separately. The overall distance is the average of these distances:

\[
D_{\text{multi}}(x, y) = \frac{1}{c} \sum_{j=1}^{c} D(x_j, y_j)
\]

where \(D(x_j, y_j)\) is the distance measure (absolute or Euclidean) between the \(j\)-th channel of the input and neuron, ensuring equal contribution from all channels.

\subsection{Prediction Algorithm}

The prediction algorithm computes distances between the input and all neurons, selecting the closest one by either averaging distances per channel or using a channel voting scheme. In channel voting, each input channel's closest neuron is identified, resulting in multiple predictions. The final prediction is the most frequent (mode) among these, or, if no consensus, the average method is used:

\[
P(x) = 
\begin{cases}
\text{mode}(\{ \arg\min_{n} D(x_j, y_{n,j}) \}), & \text{if consensus} \\
\arg\min_{n} \frac{1}{c} \sum_{j=1}^{c} D(x_j, y_{n,j}), & \text{if no consensus}
\end{cases}
\]

Where \(P(x)\) is the predicted label for input \(x\), \(D(x_j, y_{n,j})\) is the distance measure between the \(j\)-th channel of input \(x\) and the \(j\)-th channel of neuron \(n\), and \(c\) is the total number of channels.

\subsection{Implementation Details}

The implementation leverages PyTorch to process inputs in parallel, which significantly accelerates computations and enables the model to handle the MNIST and CIFAR-10 datasets efficiently. PyTorch also supports direct memory optimization, allowing the batch size to vary dynamically during parameter tuning, particularly in relation to the maximum number of neurons, which is a key factor in the algorithm's complexity. This flexibility enables us to adjust the batch size to maximize GPU memory usage, thereby processing as many inputs in parallel as possible within a single batch without exhausting memory resources.

\section{Experiments}
\subsection{Setup}
We evaluated our model on the MNIST and CIFAR-10 datasets to assess its performance on both grayscale and color image classification tasks. Each experiment was conducted five times, and the average accuracy was recorded. All the experiments were measured to take under 1 minute to complete training.

\subsection{MNIST Experiments}

\begin{table}[h!]
\centering
\begin{tabular}{|c|l|l|l|l|l|l|}
\hline
\textbf{\#} & \textbf{Description} & \textbf{Neurons} & \textbf{Distance} & \textbf{Prediction} & \textbf{Threshold}  & \textbf{Acc (\%)} \\ \hline
1 & Baseline&10 & Euclidian & Average  & False & $82.03 \pm 0.11$ \\ 
2 & Tuned & 15,000 & Euclidian & Average & False & $\textbf{97.00}\pm 0.06$ \\ \hline
\end{tabular}
\caption{Summary of Experimental Results for MNIST.}

\label{tab:results_mnist}
\end{table}

We conducted systematic experiments starting with a basic network designed for single-channel input, initially using 10 neurons, achieving an accuracy of 82.03\% (Experiment 1). This served as our baseline. After parameter tuning, increasing the neuron count to 15,000 resulted in 97\% accuracy (Experiment 2). We explored up to 30,000 neurons, beyond which the model risked memorizing the dataset.

\subsection{CIFAR-10 Experiments}
\begin{table}[h!]
\centering
\begin{tabular}{|c|c|l|l|l|l|l|l|}
\hline
{\#} & \makecell{\textbf{Description}} & \makecell{\textbf{Neu-}\\\textbf{rons}} & \makecell{\textbf{Dis-}\\\textbf{tance}} & \makecell{\textbf{Pred-}\\\textbf{iction}} & \makecell{\textbf{Thres-}\\\textbf{hold}} & \makecell{\textbf{Chan-}\\\textbf{nels}} & \makecell{\textbf{Accuracy}\\ \textbf{(\%)}} \\ \hline
1 & 1 channel & 10 & Eucl & Avg & False & 1 & $24.12\pm0.17$ \\
2 & 3 channels & 10 & Eucl & Avg & False & 3 & $27.74\pm0.24$ \\
3 & Tuning & 30,000 & Eucl & Avg & False & 3 & $36.09\pm0.45$ \\ \hline 
4 & Euclidian+Voting & 30,000 & Eucl & Voting & False & 3 & \textbf{40.60} $\pm 0.47$ \\ 
5 & Manhattan+Average & 30,000 & Manh & Avg & False & 3 & $40.25\pm0.57$ \\ 
6 & Manhattan+Voting & 30,000 & Manh & Voting & False & 3 & $40.38\pm0.43$ \\ \hline
7 & Threshold Tuning & 20,000 & Manh & Voting & True & 3 & $40.04\pm0.33$ \\
8 & Non-Threshold & 20,000 & Manh & Voting & False & 3 & $39.69\pm0.48$ \\ \hline
9 & Only channel 1 & 20,000 & Manh & Voting & True & 1 & $32.12\pm0.40$ \\ 
10 & Only channel 2 & 20,000 & Manh & Voting & True & 1 & $34.13\pm0.28$ \\ 
11 & Only channel 3 & 20,000 & Manh & Voting & True & 1 & $36.12\pm0.41$ \\ \hline
12 & Pruning & 20,000 & Manh & Voting & True & 3 & $37.88 \pm 0.66$ \\ \hline
13 & Semi-supervised & 30,000 & Eucl & Voting & False & 3 & $40.38\pm0.42$ \\ \hline
\end{tabular}
\caption{Summary of Experimental Results for CIFAR-10:  For each experiment we include a description of the experiment, the number of neurons utilized, the distance calculation method (either Euclidean or Manhattan), the prediction approach (either Average or Voting), whether a threshold was applied, the number of channels used, and the accuracy achieved, presented alongside its error margin (standard deviation).}
\label{tab:results_cifar10}
\end{table}
\subsubsection{Baseline}
Using the same model from MNIST, we initially reduced CIFAR-10's three channels to one and tested with 10 neurons, obtaining 24.12\% accuracy (Experiment 1). We then adapted the model to handle three channels directly, maintaining 10 neurons, achieving 27.74\% accuracy (Experiment 2). Parameter tuning yielded a peak accuracy of 36.09\% with 30,000 neurons, establishing our baseline (Experiment 3).

\textbf{Distance and Prediction Techniques}: Using Euclidean distance with the channel voting prediction scheme explained in section 3.7 with 30,000 neurons improved accuracy to 40.60\% (Experiment 4). An ablation study revealed that this adjustment had no impact during training, thus applied only during prediction. Switching to Manhattan distance with a classical averaging approach achieved 40.25\% (Experiment 5), while using the voting scheme reached 40.38\% (Experiment 6).

\subsubsection{Semi-supervised Learning}
We implemented a variant of the network achieving 40.60\% accuracy by initially using labels to populate neurons, once the memory is full we ran unsupervised training by assigning input to the closest neurons regardless of label. This approach resulted in a slightly lower accuracy of 40.32\% (Experiment 13).

\subsubsection{Thresholding Technique}
\begin{figure}[h!]
    \centering
    \includegraphics[width=\textwidth]{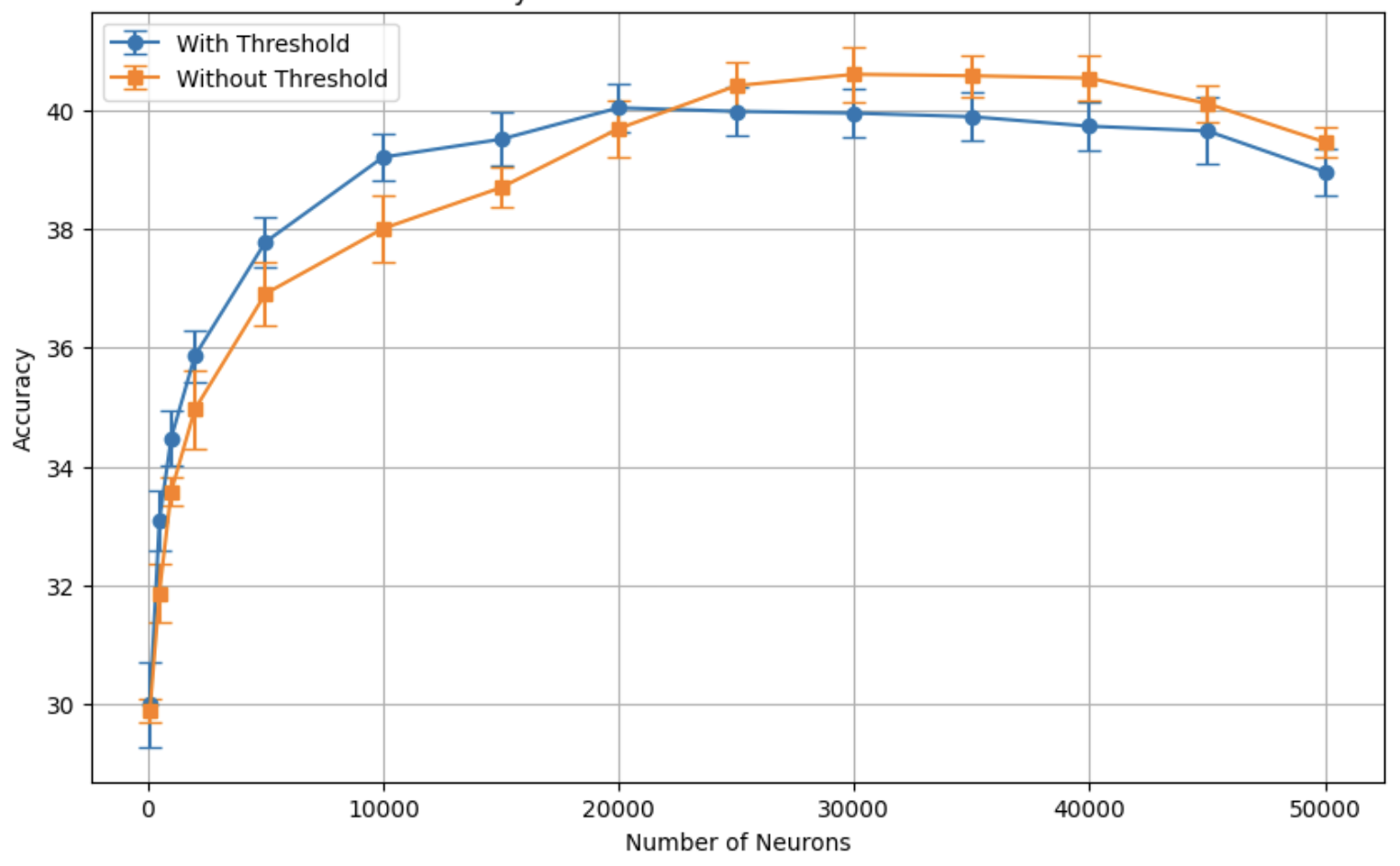} 
    \caption{Accuracy vs. Number of Neurons for two models: one using a threshold and the other without. Fourteen different values for max\_neurons were tested, with each configuration run five times. The error bars represent the standard deviation of accuracy. This figure corresponds to Experiments 7 and 8 from Table \ref{tab:results_cifar10}, illustrating that the accuracy plateaus as the number of neurons increases for both models.}
    \label{fig:accuracy_vs_neurons}
\end{figure}

Applying thresholding and tuning the model led to a performance of 40.04\% with the optimal hyperparameters being  Manhattan distance,  prediction with voting, 20,000 neurons and the coefficient 1.1 for the three channels (Experiment 7). Without thresholding, the model performed slightly lower for the same number of neurons at 39.69\% (Experiment 8).
The full parameter tuning can be visually observed in figure \ref{fig:accuracy_vs_neurons}.

\subsubsection{Other Techniques Tested}
To enhance performance with thresholding, we sorted batches by input difficulty. Input difficulty was determined by calculating the average distance between an input and other inputs of the same class across the entire training set. Reducing each batch to the easiest 20 samples resulted in 28.83\% accuracy, while using the top 50 achieved 34.2\%. Sequential training phases improved performance further, with top-to-bottom dataset transitions achieving up to 38.6\%. Sorting all inputs within batches by difficulty provided a final accuracy of 39.7\%\vspace{0.2cm}

Channel-wise classification experiments, using individual channels (and muting the other ones), resulted in accuracies of 32.12\%, 34.13\%, and 36.12\% for channels 1, 2, and 3, respectively. Muting only channel 3 achieved 36.55\%.\vspace{0.2cm}

We applied pruning to the network using a validation set consisting of 10,000 samples drawn from the original 50,000 training samples and evaluated the results on a separate test set of 10,000 samples. Previously, with thresholding, we achieved an accuracy of 40.0\%. We then tested various pruning strategies on this baseline and found the best performance by pruning 99\% of the neurons that produced errors every 100 batches. This approach resulted in an accuracy of 37.88\% on the test set and 41.26\% on the validation set.

\section{Discussion}
Our experiments provide several important insights into the model’s performance and areas for improvement. Switching from Euclidean to Manhattan distance improved accuracy (but only when using a threshold) and reduced computational complexity by simplifying distance calculations to absolute values. The choice of prediction technique, particularly the voting scheme, also enhanced accuracy by reducing the impact of outliers. While refining prediction methods might further improve performance, this was not identified as the primary limitation. Interestingly, the algorithm performed similarly in both fully supervised and semi-supervised settings, aligning with our goal of maintaining biological plausibility. A key finding, as seen in Figure \ref{fig:accuracy_vs_neurons}, is that memorization alone does not lead to optimal performance. During parameter tuning, achieving higher accuracy on CIFAR-10 required more neurons than MNIST. However, using too many neurons proved counterproductive, indicating the need for balanced model complexity. The threshold model initially outperformed the non-threshold model with fewer neurons, but as the neuron count increased, the non-threshold model ultimately achieved higher accuracy, illustrating the varying performance dynamics of both models.

The nature of the datasets also influenced the results: MNIST, with its lower variance and simpler features, allowed for higher accuracy, while CIFAR-10’s greater feature complexity made separability and representation learning more challenging. This suggests the need for techniques that reduce class variance in a biologically plausible way. Future work could explore adapting Principal Component Analysis (PCA) into a more biologically plausible preprocessing step.

\section{Conclusion}
This study introduces the Memory Network as an initial step toward developing biologically plausible alternatives to traditional deep learning models for visual classification tasks. Our experiments on the MNIST and CIFAR-10 datasets highlight both the potential and the current limitations of this approach. While the Memory Network shows promise on simpler tasks, its performance on more complex datasets like CIFAR-10 reveals the need for further refinement to better handle data variance and complexity. Integrating additional biologically plausible strategies could help address these challenges and improve overall performance.

This paper proposes a shift in research focus toward developing machine learning systems that prioritize biological plausibility in their design. Future work will explore not only enhancing existing techniques, such as Hebbian learning-based networks, to make them fully biologically plausible, but also developing entirely new methods or redesigning algorithms to better mimic the efficiency of the brain. By continuing to draw inspiration from biological mechanisms, we aim to advance toward more efficient, adaptive, and robust AI systems.

\bibliographystyle{plain}
\bibliography{bibliography}

\end{document}